\documentclass[runningheads]{llncs}

\usepackage{eccv}

\usepackage{eccvabbrv}

\usepackage{graphicx}
\usepackage{booktabs}
\usepackage{subcaption}
\usepackage{wrapfig}
\usepackage[accsupp]{axessibility}  

\usepackage{hyperref}
\usepackage{amsmath,amssymb}
\usepackage{algorithm}
\usepackage{algpseudocode} 
\usepackage{orcidlink}
\usepackage{graphicx}
\usepackage{multirow}
\usepackage{makecell}
\usepackage[table]{xcolor} 

\newcommand{\methodName}{CoverPrune}

\begin{document}

\title{CoverPrune: Coverage-Driven Token Pruning for 3D VLMs via Optimal Transport} 

\titlerunning{CoverPrune}

\author{
Peng Ling\inst{1} \and
Yingda Yin\inst{2}\textsuperscript{\dag*} \and
Lingting Zhu\inst{2}\textsuperscript{*} \and
Weikai Chen\inst{2} \and
Shengju Qian\inst{2} \and
Zeyu Hu\inst{2} \and
Xin Wang\inst{2} \and
Wenming Yang\inst{1}\textsuperscript{\dag}
}

\authorrunning{P. Ling et al.}

\institute{Shenzhen International Graduate School, Tsinghua University \and
LIGHTSPEED}

\maketitle
\begingroup
\renewcommand\thefootnote{\dag}
\footnotetext{Corresponding authors.}
\endgroup

\begingroup
\renewcommand\thefootnote{*}
\footnotetext{Project leads.}
\endgroup
\begin{abstract}
While 3D Vision-Language Models (3D VLMs) have demonstrated remarkable spatial reasoning capabilities, they suffer from massive visual token counts that create severe computational bottlenecks during inference. Existing token pruning methods primarily rely on di\-ver\-si\-ty-\allowbreak{}based selection, discarding similar tokens to maximize dispersion. However, in 3D environments, this approach frequently drops representative prototype tokens in favor of outliers, breaking the multi-view consistencies and geometric structures essential for spatial reasoning. In this paper, we propose a paradigm shift for 3D VLM token pruning: from maximizing diversity to preserving visual evidence \textit{coverage}. We introduce \methodName{}, a training-free framework that formulates inference-time token pruning as an Optimal Transport (OT) problem. To overcome the intractable combinatorial subset selection inherent in this formulation, we design the Feature-Spatial-Temporal (FST) transport cost and target capacity, along with an efficient Spatial-Guided Greedy Selection (SGS) algorithm to approximate the OT objective. Furthermore, we propose CoverPrune-Lite, an accelerated variant utilizing spatially structured local matching for minimal overhead. Extensive experiments across multiple 3D visual-spatial reasoning benchmarks demonstrate that our methods achieve state-of-the-art token efficiency, maintaining robust reasoning performance even under highly aggressive pruning budgets. Visit our project website at \url{https://github.com/Brucess/CoverPrune}.
\end{abstract}
\section{Introduction} 
Building visual-spatial intelligence with large vision-language models~\cite{bai2023qwen,lin2024video,wang2024qwen2,comanici2025gemini} has recently led to the emergence of 3D vision-language models (3D VLMs), which inject explicit geometric cues from videos, multi-view observations, or 3D representations into the visual token stream. 
While this design enables powerful spatial reasoning capabilities, it also dramatically increases the number of visual tokens processed by the model. 
A single input may generate hundreds or even thousands of tokens, causing inference to be dominated by the quadratic complexity of attention and the growth of KV caches. 
As 3D VLMs scale toward richer visual environments, visual token efficiency becomes a key bottleneck, making inference-time token pruning essential for practical deployment.
\begin{wrapfigure}{r}{0.45\textwidth}
  \centering
  \includegraphics[width=\linewidth]{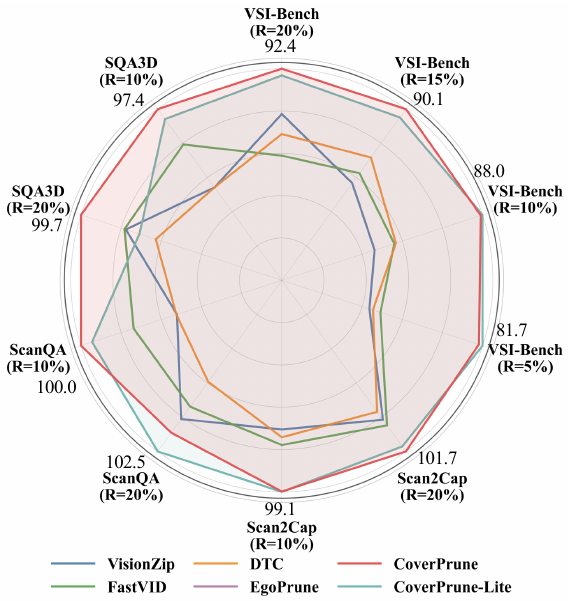}
  \caption{CoverPrune and CoverPrune-Lite Performance. We report cross-benchmark quantitative results under varying token retention ratios, with each dimension representing the average performance retention rate relative to the full token baseline across all metrics for each benchmark; our method achieves near-zero performance loss with 10\% visual tokens on general 3D tasks and retains over 90\% performance with 15\% visual tokens on the reasoning-heavy VSI-Bench.}
  \label{fig:teaser}
\end{wrapfigure}

Existing inference-time pruning methods for accelerating VLMs are primarily designed for 2D image or video understanding, and are typically evaluated on benchmarks~\cite{mme,mmbench,mvbench,longvideobench} that emphasize coarse semantic or event-level comprehension.
In these settings, token pruning is commonly performed using either \textbf{attention-based ranking} or \textbf{diversity-based selection}. 
Attention-based approaches~\cite{visionzip,sparsevlm,fastv,Ivtp} remove tokens with low early-layer attention mass, while diversity-based approaches~\cite{divprune,fastvid} treat similarity in feature space as a proxy for redundancy and remove the most mutually similar tokens to maximize dispersion among the retained set.
While these strategies are effective at reducing redundancy, they are {not explicitly designed to preserve the representative visual evidence} required for reasoning. 
Attention scores can be distorted by attention sinks and prompt-dependent saliency, while diversity-based pruning optimizes dispersion rather than representativeness.

These observations raise a fundamental question: \emph{what is the appropriate objective for token pruning in 3D VLMs?}
Existing methods\cite{tosa,dtc,egoprune} largely adopt a diversity-based perspective, where tokens that are similar to others are treated as redundant and removed to maximize dispersion among the retained set.
While this strategy can reduce redundancy when compression is mild, it becomes increasingly misaligned with representativeness under aggressive pruning.
Prototype tokens that represent dominant visual patterns are, by definition, similar to many other tokens within their mode and are therefore prone to early removal, causing the retained tokens to remain diverse yet skew toward outliers rather than representative observations.
This failure mode is particularly problematic in \textbf{3D visual-spatial reasoning}, where tokens frequently encode repeated multi-view observations that collectively establish geometric structure.
Although these observations may appear redundant in feature space, they provide complementary evidence for reconstructing spatial relationships and maintaining geometric consistency. 
Removing them based solely on similarity can therefore break multi-view correspondences and degrade reasoning about spatial relations such as distance and ordering.

Our key insight is that effective token pruning should \textbf{preserve \textit{coverage} rather than maximize diversity}. Instead of selecting tokens that are maximally different from one another, the retained tokens should collectively cover the informative content of the original token set. 
At the same time, the retained tokens must remain compact to enable efficient inference.
This principle naturally leads to selecting  {a compact set of tokens that maximizes coverage of visual evidence}.
Driven by this insight, we propose \textbf{\methodName{}}, a training-free token pruning method for 3D visual-language models that formulates pruning as a \textbf{coverage-with-compactness optimization problem}.
Rather than maximizing diversity among retained tokens, \methodName{} selects a compact subset of prototype tokens that collectively cover the informative content of the original visual token set.
Towards this end, we reinterpret token pruning through the lens of \textbf{Optimal Transport (OT)}: the retained tokens act as prototypes that distribute their representational mass to the original tokens, and pruning aims to minimize the distortion of this coverage assignment. This perspective naturally aligns token pruning with the objective of preserving representative observations while maintaining a compact token budget.

However, translating this coverage objective into a practical inference-time pruning algorithm presents several challenges. 
First, the transport cost must capture 3D representativeness, modeling feature similarity together with spatial and temporal consistency to preserve geometric structure. 
Second, under tight pruning budgets, naive transport formulations may allocate mass unevenly, requiring mechanisms that account for token informativeness. 
Third, unlike classical OT where the source and target supports are fixed, our setting requires jointly selecting prototype tokens and optimizing their coverage assignment, resulting in a combinatorial subset selection problem.

\methodName{} addresses these challenges with three key designs. 
We introduce a Feature-Spatial-Temporal (FST) transport cost that jointly models semantic similarity, spatial proximity, and temporal coherence. 
We further incorporate informativeness-aware target capacities to stabilize coverage under aggressive pruning. 
Finally, we develop an efficient Spatial-Guided Greedy Selection (SGS) algorithm that approximates the semi-relaxed OT objective for practical inference-time token selection. 
We also derive a lightweight variant, \textbf{CoverPrune-Lite}, which approximates the OT coverage objective through spatially structured local matching for faster pruning with minimal performance loss.

Extensive experiments on multiple 3D visual-spatial reasoning benchmarks demonstrate that \methodName{} consistently improves reasoning performance over state-of-the-art (SOTA) methods, as shown in Fig.~\ref{fig:teaser}. 
Compared with existing pruning methods, \methodName{} achieves better accuracy under the same token budget and maintains strong performance even under aggressive pruning. 
In summary, our main contributions are as follows:
\begin{itemize}
    \item \textbf{Novel Pruning Paradigm:} We introduce \methodName{}, a training-free token pruning framework for 3D VLMs that fundamentally shifts the pruning objective from maximizing token diversity to preserving visual evidence coverage. We elegantly formulate this via Optimal Transport (OT) to retain a compact yet highly representative set of visual tokens.
    \item \textbf{Tailored OT Solutions:} To resolve the inherent challenges of applying OT to inference-time token pruning, we propose three key designs: a Feature-Spatial-Temporal (FST) cost that comprehensively models multidimensional token relationships, dynamic target capacities that prioritize informative tokens, and an efficient optimization algorithm.
    \item \textbf{Lightweight Acceleration:} We design \textbf{CoverPrune-Lite}, a highly efficient variant that approximates the OT coverage objective through spatially structured local matching, significantly reducing pruning overhead with minimal performance degradation.
    \item \textbf{State-of-the-Art Performance:} Extensive experiments across multiple 3D visual--spatial reasoning benchmarks demonstrate that \methodName{} outperforms existing state-of-the-art token pruning methods, establishing superior token robustness, especially under aggressive pruning budgets.
\end{itemize}
\section{Related Work}
\subsection{Large Vision Language Models for 3D Understanding}
Spatial understanding is increasingly framed as a key ingredient of multimodal intelligence for embodied agents and scene centric reasoning \cite{thinking,videorft,lee2025perspective,gpt4scene,ling2025sovgaussian}. Recent VLMs provide a strong base by pairing robust 2D perception with scalable language reasoning \cite{bai2023qwen,lin2024video,wang2024qwen2,comanici2025gemini}. Building on these pretrained backbones, a dominant paradigm introduces explicit geometry into visual tokens so that spatial reasoning can rely on 3D evidence rather than implicitly recovering structure from 2D cues. A series of works inject 3D signals derived from SfM or geometry foundation models into pre-trained LVLMs, and finetune the resulting 3D VLMs for spatial question answering tasks. SR-3D \cite{sr3d} incorporates 3D aware region representations to support spatially grounded language interaction. Spatial-MLLM \cite{spatialmllm} and GS-Reasoner \cite{gsreasoner} further emphasize geometry augmented tokenization and grounded reasoning, showing that explicit structure improves spatial queries under viewpoint changes. Other efforts broaden supervision and objectives for 3D and video spatial understanding, including position aware training signals and instruction aligned spatial tuning \cite{learningfor3d,vlm3r}. Despite rapid progress in 3D VLMs, the critical issue of token explosion induced by multi-frame inputs has yet to be adequately addressed, leaving efficient inference a long-standing bottleneck for them with spatial reasoning capabilities.
\begin{figure}[tb]
  \centering
  \includegraphics[width=0.95\linewidth]{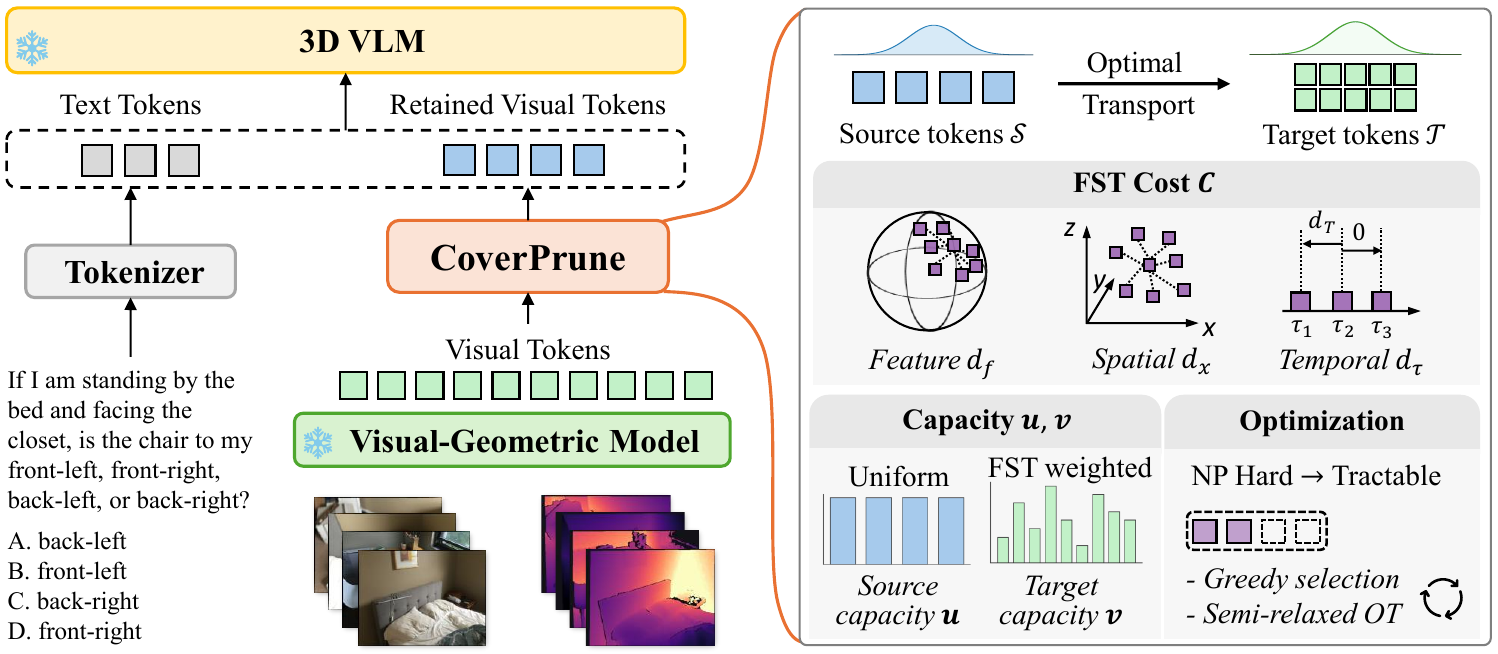}
  \caption{\textbf{Framework overview.} 
    \textbf{(Left)} \methodName{} serves as a training-free, plug-and-play module inserted between the visual-geometric encoder and the 3D VLM. 
    \textbf{(Right)} We formulate token pruning as an Optimal Transport (OT) problem to maximize visual evidence coverage. 
    To resolve this, we introduce three key designs: (1) a Feature-Spatial-Temporal (FST) Cost $\mathbf{C}$ ($d_f$, $d_x$, $d_\tau$) to comprehensively model multidimensional token relationships; (2) an asymmetric capacity assignment to stabilize mass allocation based on token informativeness; and (3) a tractable optimization strategy to approximate the inherently NP-hard combinatorial subset selection problem.}
  \label{fig:figure1}
\end{figure}

\subsection{Visual Token Pruning}
Visual token reduction is widely studied for accelerating VLM inference in generic image and video understanding, where evaluation typically emphasizes coarse perception and caption style or short form reasoning rather than quantitative spatial grounding \cite{mme,mmbench,mvbench,longvideobench}. One direction compresses tokens through redesigned multimodal projectors. Honeybee \cite{honeybee}, LLaVA-UHD \cite{llavauhd}, and TokenPacker \cite{tokenpacker} reduce visual token counts before the language model, but commonly require architectural changes and end-to-end adaptation. A lighter line performs training-free pruning and mainly follows attention-based or diversity-based criteria\cite{empirical}. Attention-based approaches \cite{visionzip,sparsevlm,fastv,Ivtp} estimate token saliency from self attention or cross-modal attention and remove low score tokens. Diversity-based methods \cite{divprune,fastvid} reduce redundancy by feature space merging or maximizing diversity of retained set. While effective for generic understanding, these heuristics can be misaligned with spatial reasoning, since attention patterns can evolve across layers and decoding steps, and feature-driven merging can distort the representative tokens needed for complex spatial reasoning. Only a few works explicitly tailor token reduction to spatial tasks. DTC \cite{dtc} compresses inputs for 3D question answering with voxel grounded token compression, EgoPrune \cite{egoprune} leverages SfM pose cues to align overlapping regions before filtering redundant tokens, and ToSA \cite{tosa} introduces spatial awareness signals to guide safer merging.

\section{CoverPrune}
\subsection{Preliminary: Optimal Transport}
Optimal Transport (OT) \cite{optimaltransport} provides a principled way to measure how well one weighted set can be matched to another under a chosen notion of cost.
Consider a source set $\{s_i\}_{i=1}^{m}$ and a target set $\{t_j\}_{j=1}^{n}$, equipped with nonnegative capacities
$\mathbf{u}\in\mathbb{R}^m_{+}$ and $\mathbf{v}\in\mathbb{R}^n_{+}$, which are often normalized so $\mathbf{u}^\top\mathbf{1}=1$ and $\mathbf{v}^\top\mathbf{1}=1$.
Throughout this paper, we use capacity to refer to these OT marginal vectors (i.e., distributions and weights).
Let $\mathbf{C}\in\mathbb{R}^{m\times n}$ be a cost matrix, where $C_{ij}$ quantifies the cost of assigning $s_i$ to $t_j$.
A transport plan is a nonnegative matrix $\mathbf{P}\in\mathbb{R}^{m\times n}_{+}$ whose row and column sums match the prescribed capacities:
\begin{equation}
\mathbf{P}\mathbf{1}=\mathbf{u},\qquad \mathbf{P}^\top\mathbf{1}=\mathbf{v}.
\end{equation}
The OT objective finds the least-cost plan:
\begin{equation}
\label{eq:ot}
\mathrm{OT}(\mathbf{u},\mathbf{v})=\min_{\mathbf{P}\ge 0}\ \langle \mathbf{C},\mathbf{P}\rangle 
\ \ \text{s.t.}\ \ \mathbf{P}\mathbf{1}=\mathbf{u},\ \mathbf{P}^\top\mathbf{1}=\mathbf{v},
\end{equation}
where $\langle \mathbf{C},\mathbf{P}\rangle=\sum_{i,j}C_{ij}P_{ij}$.

For efficiency, a common practice is to add an entropic regularizer with weight $\varepsilon>0$ and solve the smoothed problem with Sinkhorn \cite{sinkhorn} iterations:
\begin{equation}
\label{eq:ot_entropic}
\mathrm{OT}_{\varepsilon}(\mathbf{u},\mathbf{v})=
\min_{\mathbf{P}\ge 0}\ \langle \mathbf{C},\mathbf{P}\rangle
-\varepsilon H(\mathbf{P})
\ \ \text{s.t.}\ \ \mathbf{P}\mathbf{1}=\mathbf{u},\ \mathbf{P}^\top\mathbf{1}=\mathbf{v},
\end{equation}
where $H(\mathbf{P})=-\sum_{i,j}P_{ij}(\log P_{ij}-1)$.

\subsection{Problem Setup}
The overview of CoverPrune is shown in Fig.~\ref{fig:figure1}. We first introduce the problem formulation. Let $\mathcal{T}=\{t_j\}_{j=1}^{N}$ denote all visual tokens extracted from an input video before being fed into the backbone of a 3D VLM.
Each token $t_j$ is associated with a feature embedding $\mathbf{f}_j\in\mathbb{R}^{d}$, a 3D global coordinate $\mathbf{x}_j\in\mathbb{R}^{3}$ that can be estimated via SfM or a geometry foundation model, and a timestamp $\tau_j$ of its corresponding frame.
Given a retention ratio $R\in(0,1]$, we set the pruning budget as $K=\lceil R N\rceil$ and aim to select a subset $\mathcal{S}\subseteq\mathcal{T}$ with $|\mathcal{S}|=K$ as the visual input for subsequent decoding:
\begin{equation}
\label{eq:pruning_problem}
\mathcal{S}^{\star}
=
\arg\max_{\mathcal{S}\subseteq\mathcal{T},\ |\mathcal{S}|=K}
\ \mathrm{Cover}(\mathcal{S};\mathcal{T}).
\end{equation}

To obtain a principled and computable notion of coverage, we cast prototype selection as minimizing the discrepancy between the selected set and the full token set. Concretely, we treat $\mathcal{S}$ as a source support that should explain the target support $\mathcal{T}$, and measure their mismatch via an OT objective. We assign nonnegative capacities to tokens in $\mathcal{S}$ and $\mathcal{T}$, define a pairwise cost between any retained token and any original token, and compute an OT matching cost by optimizing a transport plan:
\begin{equation}
\label{eq:3dto_ot}
\mathcal{L}_{\mathrm{OT}}(\mathcal{S};\mathcal{T})
=
\min_{\mathbf{P}\ge 0}\ \langle \mathbf{C}(\mathcal{S},\mathcal{T}), \mathbf{P} \rangle
\quad
\text{s.t.}\quad
\mathbf{P}\mathbf{1}=\mathbf{u},\ \mathbf{P}^{\top}\mathbf{1}=\mathbf{v}.
\end{equation}
Here, $\mathbf{C}(\mathcal{S},\mathcal{T})$ is the cost matrix with entries $C_{ij}$ measuring the discrepancy between a retained token $s_i\in\mathcal{S}$ and an original token $t_j\in\mathcal{T}$; $\mathbf{P}$ denotes the transport plan; and $\mathbf{u}$ and $\mathbf{v}$ are normalized capacity vectors over $\mathcal{S}$ and $\mathcal{T}$, respectively. In our setting, choosing $\mathbf{u}$ to be uniform is a natural and reasonable default, since the selected prototypes serve as an unlabeled summary set without prior preference among them. In contrast, a uniform $\mathbf{v}$ is generally suboptimal because original tokens can vary substantially in informativeness, and treating them equally may allocate excessive mass to uninformative or noisy regions while under-emphasizing salient spatial evidence.

CoverPrune then selects the retained token set by minimizing this transport cost:
\begin{equation}
\label{eq:3dto_select}
\mathcal{S}^{\star}
=
\arg\min_{\mathcal{S}\subseteq\mathcal{T},\ |\mathcal{S}|=K}
\ \mathcal{L}_{\mathrm{OT}}(\mathcal{S};\mathcal{T}).
\end{equation}
In the following subsections, we detail how to instantiate Eq.~\eqref{eq:3dto_ot}--\eqref{eq:3dto_select} by (i) designing a multi-domain transport cost $\mathbf{C}(\mathcal{S},\mathcal{T})$, (ii) specifying the target capacities $\mathbf{v}$ via an informativeness-aware reweighting scheme, and (iii) developing an efficient inference-time optimization strategy for selecting $\mathcal{S}$ despite the underlying NP-hard combinatorial search.

\subsection{Feature-Spatial-Temporal Transport Cost} 
3D visual-spatial reasoning demands information preservation beyond pure semantics. Prior pruning methods judge redundancy solely in the feature domain, risking damage to the geometric structure critical for grounding, while ignoring temporal order introduces inconsistent evidence of region observation timing and location. To address this, we design a Feature-Spatial-Temporal (FST) criterion to jointly model token relations across these three domains: (i) feature-space proximity minimizes semantic distortion during token substitution; (ii) 3D spatial proximity preserves geometric integrity for accurate object and relation grounding in reasoning; (iii) temporal consistency aligns the retained token set with the true order of spatial evidence, mitigating errors from temporal mis-association. 

To operationalize this FST criterion, we quantify three pairwise discrepancies between a token $s$ and a token $t$: \begin{equation}\label{eq:fst_metrics}
d_f(s,t)=1-\cos(\mathbf{f}_s,\mathbf{f}_t),\quad
d_x(s,t)=\|\mathbf{x}_s-\mathbf{x}_t\|_2,\quad
d_\tau(s,t)=\mathrm{ReLU}(\tau_s-\tau_t),
\end{equation}
where $\mathbf{f}$ denotes the token embedding, $\mathbf{x}$ is the 3D global coordinate, and $\tau$ is the timestamp. The temporal term uses $\mathrm{ReLU}(z)=\max(z,0)$ to penalize covering a token observed earlier in time with one observed later, encouraging the retained set to respect the temporal order of spatial evidence. We define the transport cost between a retained token $s_i\in\mathcal{S}$ and an original token $t_j\in\mathcal{T}$ as a weighted sum:
\begin{equation}
\label{eq:fst_cost}
C_{ij}
=
\lambda_f\,\hat d_f(s_i,t_j)
+\lambda_x\,\phi_{\kappa}\!\big(\hat d_x(s_i,t_j)\big)
+\lambda_\tau\,\hat d_\tau(s_i,t_j),
\end{equation}
where $\hat d(\cdot,\cdot)$ denotes a min-max normalized discrepancy computed within the current sample, and $\lambda_f,\lambda_x,\lambda_\tau$ control the relative importance of feature, spatial, and temporal terms. We further apply a nonlinear mapping $\phi_{\kappa}(x)=\log(1+\kappa x)/\log(1+\kappa)$ with $\kappa>0$, which expands the dynamic range for near-field distances. $C_{ij}$ serves as a unified notion of substitutability. Consequently, the transport plan favors allocating mass along semantically aligned, geometrically consistent, and temporally coherent correspondences, which directly steers the selected tokens toward globally faithful coverage.

\subsection{Feature-Spatial-Temporal Capacity} The transport cost in Eq.~\eqref{eq:fst_cost} specifies how mass should be assigned once $\mathcal{S}$ is given, but effective coverage under a tight budget also depends on which target tokens deserve more coverage. We therefore introduce an FST capacity vector $\mathbf{v}\in\mathbb{R}^{N}_{+}$ to parameterize the target capacity in Eq.~\eqref{eq:3dto_ot}, where a larger $v_j$ encourages allocating more transport mass to token $t_j$. Our design follows the same FST criterion: tokens that are harder to approximate by their local neighborhood in the FST sense tend to carry more distinctive information and should be prioritized.

Specifically, for each target token $t_j$, we compute a local distinctiveness score by averaging its FST discrepancy to a small neighbor set $\mathcal{N}_{n}(t_j)$:
\begin{equation}
\label{eq:fst_capacity_score}
r_j
=
\frac{1}{|\mathcal{N}_{n}(t_j)|}\sum_{t_k\in\mathcal{N}_{n}(t_j)}
\Big(
\alpha_f\,\hat d_f(t_j,t_k)
+\alpha_x\,\hat d_x(t_j,t_k)
+\alpha_\tau\,\hat d_\tau(t_j,t_k)
\Big),
\end{equation}
where $\mathcal{N}_{n}(t_j)$ denotes the set of $n$ nearest neighbors of $t_j$ in 3D space, and $\alpha_f,\alpha_x,\alpha_\tau$ are capacity weights. We then map $\{r_j\}$ to a nonnegative capacity vector and normalize it to match the pruning budget:
\begin{equation}
\label{eq:fst_capacity}
v_j
=
\frac{\phi(r_j)}{\sum_{l=1}^{N}\phi(r_l)},
\end{equation}
where $\phi(\cdot)$ is a monotone increasing mapping that controls how capacity concentrates on informative tokens. This construction allocates more capacity to tokens that are locally distinctive in feature, geometry, or time, preventing dense but redundant regions from dominating the coverage objective.

\subsection{Optimization}
\subsubsection{Semi-Relaxed Optimal Transport.}
\label{sec:srot}
Classical OT in Eq.~\eqref{eq:ot} assumes that both supports and their weights are fixed, and optimizes only the transport plan. 
In our setting, however, the source support is itself a decision variable: we seek a subset $\mathcal{S}\subseteq \{s_i\}_{i=1}^{m}$ with $|\mathcal{S}|\le K$ and a capacity vector supported on $\mathcal{S}$ that best matches a fixed target capacity under the OT cost. This yields a coupled discrete--continuous problem, where one must jointly select $S$ and solve for the optimal coupling. Such formulations are generally intractable to solve exactly and are known to be NP-hard\cite{spot,pwc}.

To obtain an efficient solver with provable approximation behavior, an approach is to relax the strict OT marginal constraints while preserving the OT principle of minimizing transport cost. The key idea is to introduce slack on the target side so that the induced set objective becomes amenable to greedy optimization, and in particular to submodular maximization. Concretely, we adopt the semi-relaxed OT formulation \cite{benamou2015bregman,peyre2019computational,chapel2020pot}:
\begin{equation}
\label{eq:srpot}
\mathrm{SOT}(\mathbf{u},\mathbf{v})
=\min_{\mathbf{P}\ge 0}\ \langle \mathbf{C},\mathbf{P}\rangle
\quad \text{s.t.}\quad
\mathbf{P}\mathbf{1}= \mathbf{u},\ \ \mathbf{P}^\top\mathbf{1}\le \mathbf{v},
\end{equation}
where $\mathbf{u}\in\mathbb{R}^m_{+}$ denotes the source capacity and $\mathbf{v}\in\mathbb{R}^n_{+}$ specifies per-target capacities.
Compared to OT, the inequality constraint $\mathbf{P}^\top\mathbf{1}\le \mathbf{v}$ permits unused capacity, which provides exactly the flexibility needed for tractable support optimization. Prior work shows that, under such relaxed Wasserstein objectives, the induced set functions for subset selection can be monotone submodular, and therefore admit greedy maximization with constant-factor approximation guarantees under a cardinality constraint \cite{spot,pwc}. When the relaxation is tight, semi-relaxed OT recovers standard OT as a special case \cite{peyre2019computational,pwc}. Thus, semi-relaxed OT serves as a principled relaxation that enables greedy subset construction while remaining consistent with the OT objective at the target budget. Similar to Eq.~\eqref{eq:ot_entropic}, we further add an entropic regularizer to obtain a smooth objective:
\begin{equation}
\label{eq:srpot_sinkhorn}
\mathrm{SOT}_{\varepsilon}(\mathbf{u},\mathbf{v})
=\min_{\mathbf{P}\ge 0}\ \langle \mathbf{C},\mathbf{P}\rangle
-\varepsilon H(\mathbf{P})
\quad \text{s.t.}\quad
\mathbf{P}\mathbf{1}= \mathbf{u},\ \ \mathbf{P}^\top\mathbf{1}\le \mathbf{v}.
\end{equation}

\subsubsection{Spatial-Guided Greedy Selection.}
While the semi-relaxed formulation makes subset construction algorithmically feasible, a naive greedy implementation is still computationally prohibitive for long video sequences. In particular, evaluating the marginal benefit of adding each candidate token would require re-solving a transport problem at every greedy step, resulting in excessive runtime.

We therefore propose Spatial-Guided Greedy Selection (SGS), built on a simple locality prior in 3D scenes: a token can effectively cover only tokens that are spatially nearby, since distant 3D regions typically correspond to different surfaces or objects and incur high transport cost. This observation motivates restricting marginal-cost evaluation to a small 3D neighborhood, which avoids repeated global OT computations from each candidate token to all target tokens. Specifically, for each candidate token $t$, we define its neighborhood within the target set as
\begin{equation}
\label{eq:local_neighbor}
\mathcal{M}_{g}(t)=\text{NN}_{g}\!\left(t;\mathcal{T}\right),
\end{equation}
where $\text{NN}_{g}$ returns the $g$ nearest target tokens in 3D space. At greedy step $\ell$, given the current selected set $\mathcal{S}_{\ell}$, we solve a single semi-relaxed OT problem to obtain a transport plan $\mathbf{P}_{\ell}$ and compute the residual capacity on the target side:
\begin{equation}
\label{eq:greedy_flow}
\mathbf{r}_{\ell}=\Big[\mathbf{v}-\mathbf{P}_{\ell}^{\top}\mathbf{1}\Big]_{+}.
\end{equation}
We then select the next token by minimizing a local, residual-weighted marginal cost:
\begin{equation}
\label{eq:sgs_select}
t_{\ell}^{\star}
=
\arg\min_{t\in\mathcal{T}\setminus\mathcal{S}_{\ell}}
\ \sum_{t_j\in\mathcal{M}_{g}(t)} r_{\ell,j}\,C(t,t_j),
\end{equation}
and update the set as
\begin{equation}
\label{eq:sgs_update}
\mathcal{S}_{\ell+1}=\mathcal{S}_{\ell}\cup\{t_{\ell}^{\star}\},
\end{equation}
where $C(\cdot,\cdot)$ is the FST cost and $\mathbf{v}$ is the corresponding capacity vector.

\section{CoverPrune-Lite: Block-Structured OT Approximation via 3D-Aware Ordering}
CoverPrune constructs the retained set with a greedy procedure that repeatedly solves semi-relaxed OT. While principled, this iterative global transport optimization incurs cubic-time complexity in the number of tokens, which becomes a bottleneck for long video sequences. To further improve efficiency, we propose \textbf{CoverPrune-Lite}, which exploits a spatial locality prior that effective coverage is predominantly supported by spatially nearby tokens in 3D. CoverPrune-Lite therefore approximates OT-style coverage by restricting transport to local 3D neighborhoods, eliminating iterative transport solving.

\subsection{3D-Aware Ordering and Capacity-Guided Grouping}
We first build a 3D-aware ordering of all target tokens via the Morton code space-filling curve \cite{samet}. This ordering preserves spatial locality, so tokens that are adjacent in the sorted list are likely to be proximal in 3D space, enabling coherent neighborhood construction without explicit nearest-neighbor search. On top of this order, we partition tokens into $K$ non-overlapping groups using the target FST capacity. Let $\mathbf{v}\in\mathbb{R}^{N}_{+}$ denote the per-token target capacity, normalized as a distribution with $\sum_{j=1}^{N} v_j=1$. We traverse the Morton-ordered list and accumulate capacity until it reaches $1/K$, then finalize a group and start a new one, producing $K$ groups $\{\mathcal{G}_q\}_{q=1}^{K}$ that satisfy
\begin{equation}
\label{eq:lite_group}
\sum_{t_j\in \mathcal{G}_q} v_j \approx \frac{1}{K} \qquad q=1,\ldots,K.
\end{equation}
This adaptive grouping assigns approximately equal information mass to each group. Regions with many redundant tokens tend to have small per-token capacity and thus form larger groups, whereas informative regions have larger per-token capacity and form smaller, more fine-grained groups.

We then choose one prototype token from each group by minimizing the capacity-weighted transport cost within the group,
\begin{equation}
\label{eq:lite_select}
s_q
=
\arg\min_{t\in \mathcal{G}_q}\ \sum_{t_j\in \mathcal{G}_q} v_j\, C(t,t_j),
\end{equation}
and output the pruned set as $\mathcal{S}=\{s_q\}_{q=1}^{K}$.

\subsection{Connection to Coverage Objective}
CoverPrune-Lite can be understood as a block-constrained variant of our OT objective (i.e., Eq.~\eqref{eq:3dto_select}), where the transport plan is restricted to be locally supported on the pre-defined groups. With the same source and target capacities $\mathbf{u}$ and $\mathbf{v}$ defined in CoverPrune, we restrict the feasible couplings to
\begin{equation}
\label{eq:blk_plan}
\mathcal{P}_{\mathrm{blk}}
=
\Bigl\{\mathbf{P}\ge 0 \,\big|\, \mathbf{P}\mathbf{1}=\mathbf{u},\ \mathbf{P}^{\top}\mathbf{1}= \mathbf{v},\ P_{qj}=0\ \text{if}\ t_j\notin \mathcal{G}_q\Bigr\}.
\end{equation}
Here we take a uniform source capacity over the $K$ retained tokens. For $q=1,\ldots,K$, so that $\sum_q u_q = 1 = \sum_j v_j$.

Given a fixed partition $\{\mathcal{G}_q\}$, we then consider the constrained objective
\begin{equation}
\label{eq:blk_ot}
\min_{\substack{s_q\in \mathcal{G}_q\\ q=1,\ldots,K}}\ \ \min_{\mathbf{P}\in \mathcal{P}_{\mathrm{blk}}}\ \big\langle \mathbf{C}(\mathcal{S},\mathcal{T}),\mathbf{P}\big\rangle .
\end{equation}
When the grouping satisfies $\sum_{t_j\in \mathcal{G}_q} v_j = 1/K$, the block constraint forces each prototype token $s_q$ to send its entire mass $u_q=1/K$ within $\mathcal{G}_q$. In this case, the optimal coupling is uniquely determined as $P_{qj}=v_j$ for $t_j\in\mathcal{G}_q$ and $P_{qj}=0$ otherwise. Substituting this into Eq.~\eqref{eq:blk_ot} yields $\sum_{q=1}^{K}\sum_{t_j\in \mathcal{G}_q} v_j\,C(s_q,t_j)$, which decouples across groups and recovers exactly the within-group selection rule in Eq.~\eqref{eq:lite_select}. Therefore, CoverPrune-Lite approximates CoverPrune by enforcing a block-diagonal, locally supported transport structure, replacing iterative global OT optimization with a single pass of 3D-aware grouping and locally optimal prototype selection, and achieving $O(N\log N)$ inference-time complexity.
\section{Experiments}
\subsection{Experimental Settings}
\textbf{Datasets and Benchmarks. } 
To comprehensively evaluate our training-free CoverPrune and CoverPrune-Lite, we conduct extensive experiments on four mainstream 3D vision-language benchmarks. We first validate our method on three widely used fine-grained reasoning benchmarks: ScanQA \cite{scanqa}, SQA3D \cite{sqa3d}, and Scan2Cap \cite{scan2cap}. We further apply our method to VSI-Bench \cite{thinking}, an egocentric indoor scan-based video benchmark for complex spatial-temporal reasoning, with full evaluation across eight tasks: Object Count, Relative Distance, Relative Direction, Route Plan, Object Size, Room Size, Absolute Distance, and Appearance Order.

\subsubsection{Baselines.}
We compare CoverPrune and CoverPrune-Lite against four state-of-the-art (SOTA) training-free visual token pruning methods: two general multimodal methods, VisionZip \cite{visionzip} and FastVID \cite{fastvid}, which integrate attention-based importance estimation with feature diversity heuristics; two 3D VLM-specific methods, DTC \cite{dtc} with a diversity-driven token selection strategy, and EgoPrune \cite{egoprune} with attention-diversity fused token merging.

\subsubsection{Implementation Details.}
To validate the generalizability of our proposed method, we instantiate it on two SOTA 3D VLMs, GS-Reasoner \cite{gsreasoner} and VLM-3R \cite{vlm3r}, both augmenting visual tokens with geometric cues for 3D spatial reasoning. We keep all default base-model configurations fully unchanged for a fair, controlled comparison, including uniform 32-frame sampling, and follow the protocol in GS-Reasoner to generate all token coordinates via an estimator without using ground-truth values. Our methods are inserted immediately before the LLM prefill stage, operating directly on raw visual tokens with compatibility across diverse acceleration frameworks. We set $\lambda_f=\lambda_x=\lambda_\tau=1$ and $\alpha_f=\alpha_x=\alpha_\tau=1$ in our experiments. 

\begin{table}[tb]
  \caption{Evaluation on General 3D Tasks. Vanilla baseline results from GS-Reasoner~\cite{gsreasoner}. Retention ratio $R$ is the fraction of visual tokens retained post-pruning. Per-benchmark Acc.\% is the average relative performance retention across all its metrics. CoverPrune and its Lite variant consistently hit top-1/top-2 on most metrics, with strong multi-task robustness.}
  \label{tab:scan_tasks}
  \centering
  \small
  \setlength{\tabcolsep}{4pt}
  \renewcommand{\arraystretch}{1.15}

  \resizebox{\linewidth}{!}{%
  \begin{tabular}{@{}l c ccccc cccccc cc@{}}
    \toprule
    \multirow{2}{*}{Method} &
    \multirow{2}{*}{\makecell{Retention\\Ratio $R$}} &
    \multicolumn{5}{c}{Scan2Cap} &
    \multicolumn{6}{c}{ScanQA} &
    \multicolumn{2}{c}{SQA3D} \\
    \cmidrule(lr){3-7}\cmidrule(lr){8-13}\cmidrule(lr){14-15}
     & &
     Acc.\%$\uparrow$ & B-4$\uparrow$ & Rouge$\uparrow$ & CIDEr$\uparrow$ & Meteor$\uparrow$ &
     Acc.\%$\uparrow$ & B-4$\uparrow$ & Rouge$\uparrow$ & CIDEr$\uparrow$ & Meteor$\uparrow$ & EM$\uparrow$ &
     Acc.\%$\uparrow$ & EM$\uparrow$ \\
    \midrule

    Vanilla & 100\% &
      100.00 & 47.60 & 69.20 & 101.00 & 32.10 &
      100.00 & 16.20 & 49.20 & 102.60 & 19.80 & 29.90 &
      100.00 & 59.90 \\
    \midrule

    VisionZip {\scriptsize (CVPR25)}       & \multirow{6}{*}{20\%} &
      99.29 & 49.02 & 70.96 & 93.48 & 31.81 &
      99.85 & \cellcolor{red!12}17.68 & 47.77 & 101.19 & 19.71 & 28.36 &
      97.96 & 58.68 \\
    FastVID {\scriptsize (NeurIPS25)}      &                        &
      99.71 & 49.15 & 70.85 & 94.78 & 31.90 &
      98.85 & 16.60 & 48.17 & 101.30 & 19.73 & 28.56 &
      \cellcolor{yellow!12}98.01 & \cellcolor{yellow!12}58.71 \\
    DTC {\scriptsize (CVPR25)}             &                        &
      98.70 & 48.54 & 70.52 & 93.04 & 31.71 &
      96.85 & 16.14 & 47.38 & 98.80  & 19.29 & 28.28 &
      96.83 & 58.00 \\
    EgoPrune {\scriptsize (arXiv25)}       &                        &
      88.79 & 43.95 & 68.49 & 71.66 & 29.82 &
      88.67 & 15.28 & 43.83 & 88.55  & 17.86 & 24.94 &
      92.04 & 55.13 \\
    CoverPrune                              &                        &
      \cellcolor{red!12}101.68 & \cellcolor{red!12}50.10 & \cellcolor{red!12}71.18 & \cellcolor{red!12}99.35 & \cellcolor{red!12}32.18 &
      \cellcolor{yellow!12}100.95 & 16.99 & \cellcolor{yellow!12}48.87 & \cellcolor{yellow!12}103.17 & \cellcolor{yellow!12}20.04 & \cellcolor{red!12}29.54 &
      \cellcolor{red!12}99.67 & \cellcolor{red!12}59.70 \\
    CoverPrune-Lite                         &                        &
      \cellcolor{yellow!12}101.31 & \cellcolor{yellow!12}49.94 & \cellcolor{yellow!12}71.17 & \cellcolor{yellow!12}98.22 & \cellcolor{yellow!12}32.17 &
      \cellcolor{red!12}102.46 & \cellcolor{yellow!12}17.63 & \cellcolor{red!12}49.42 & \cellcolor{red!12}104.69 & \cellcolor{red!12}20.32 & \cellcolor{yellow!12}29.41 &
      97.45 & 58.37 \\
    \midrule

    VisionZip {\scriptsize (CVPR25)}       & \multirow{6}{*}{10\%} &
      93.65 & 46.61 & 69.62 & 81.30 & 30.68 &
      90.76 & 15.23 & 44.66 & 92.13  & 18.29 & 25.97 &
      92.60 & 55.47 \\
    FastVID {\scriptsize (NeurIPS25)}      &                        &
      95.00 & 46.62 & 69.75 & 85.68 & 30.96 &
      94.93 & 15.58 & 46.67 & 97.51  & 19.04 & 27.64 &
      95.26 & 57.06 \\
    DTC {\scriptsize (CVPR25)}             &                        &
      94.33 & 46.60 & 69.66 & 83.42 & 30.87 &
      90.81 & 14.84 & 45.01 & 92.50  & 18.24 & 26.52 &
      92.60 & 55.47 \\
    EgoPrune {\scriptsize (arXiv25)}       &                        &
      80.69 & 40.23 & 66.73 & 53.93 & 28.38 &
      80.66 & 13.66 & 40.39 & 80.01  & 16.42 & 22.72 &
      86.86 & 52.03 \\
    \textbf{CoverPrune}                     &                        &
      \cellcolor{yellow!12}99.04 & \cellcolor{red!12}48.85 & \cellcolor{yellow!12}70.63 & \cellcolor{red!12}93.84 & \cellcolor{yellow!12}31.64 &
      \cellcolor{red!12}100.00 & \cellcolor{red!12}17.63 & \cellcolor{red!12}47.98 & \cellcolor{red!12}100.87 & \cellcolor{red!12}19.71 & \cellcolor{red!12}28.64 &
      \cellcolor{red!12}97.45 & \cellcolor{red!12}58.37 \\
    \textbf{CoverPrune-Lite}                &                        &
      \cellcolor{red!12}99.07 & \cellcolor{yellow!12}48.75 & \cellcolor{red!12}70.69 & \cellcolor{yellow!12}93.71 & \cellcolor{red!12}31.76 &
      \cellcolor{yellow!12}98.93 & \cellcolor{yellow!12}17.34 & \cellcolor{yellow!12}47.66 & \cellcolor{yellow!12}100.29 & \cellcolor{yellow!12}19.55 & \cellcolor{yellow!12}28.19 &
      \cellcolor{yellow!12}96.83 & \cellcolor{yellow!12}58.00 \\
    \bottomrule
  \end{tabular}%
  }
\end{table}
\subsection{Effectiveness Evaluation}
\subsubsection{Results on General 3D Tasks.}
Table~\ref{tab:scan_tasks} demonstrates that CoverPrune and CoverPrune-Lite achieve top-tier performance on nearly all reported metrics under matched token budgets, consistently ranking first or tied for first across 3D captioning and question answering tasks with well-generalized performance gains. Critically, our performance lead over competing baselines widens significantly under more aggressive pruning. However, we observe that SOTA performance on general 3D QA tasks has largely saturated, motivating us to pioneer systematic token pruning evaluation on VSI-Bench, a visual-spatial reasoning benchmark with substantially higher complexity and reasoning difficulty.
\begin{table}[tb]
  \caption{Evaluation on VSI-Bench with GS-Reasoner as the base model. Under each ratio, the number of tokens retained by each method was kept consistent. Our proposed CoverPrune and CoverPrune-Lite consistently exhibit superior performance.}
  \label{tab:qa_style_table_gs}
  \centering
  \small
  \setlength{\tabcolsep}{4pt}
  \renewcommand{\arraystretch}{1.15}
  \resizebox{\linewidth}{!}{%
  \begin{tabular}{@{}l c c cccccccc@{}}
    \toprule
    \multirow{2}{*}{Method} &
    \multirow{2}{*}{\makecell{Retention\\Ratio $R$}} &
    \multirow{2}{*}{\makecell{Avg.}} &
    \multicolumn{4}{c}{\cellcolor{orange!12}\textbf{Numerical Answer}} &
    \multicolumn{4}{c}{\cellcolor{yellow!12}\textbf{Multiple-Choice Answer}} \\
    & & &
    Obj.\ Count &
    Abs.\ Dist. &
    Obj.\ Size &
    Room Size &
    Rel.\ Dist. &
    Rel.\ Dir. &
    Route Plan &
    Appr.\ Order \\
    \midrule

    Vanilla & 100\% & 64.70 & 69.10 & 61.90 & 70.00 & 65.70 & 65.40 & 88.90 & 44.30 & 52.30 \\
    \midrule

    VisionZip {\scriptsize (CVPR25)}       & \multirow{6}{*}{20\%} & 57.55 & 67.82 & 50.16 & 61.10 & 53.26 & \cellcolor{red!12}59.86 & 78.31 & \cellcolor{red!12}40.72 & 49.19 \\
    FastVID {\scriptsize (NeurIPS25)}      &                        & 55.52 & 67.04 & 49.62 & 63.85 & 52.81 & 54.65 & 73.40 & 34.54 & 48.22 \\
    DTC {\scriptsize (CVPR25)}             &                        & 56.57 & 68.55 & 51.03 & 64.80 & 46.91 & 55.07 & 78.99 & 35.57 & 51.62 \\
    EgoPrune {\scriptsize (arXiv25)}       &                        & 49.44 & 63.72 & 43.53 & 59.34 & \cellcolor{yellow!12}54.06 & 44.51 & 65.84 & 35.57 & 28.96 \\
    \textbf{CoverPrune}                             &                        & \cellcolor{red!12}59.76 & \cellcolor{red!12}69.38 & \cellcolor{red!12}55.04 & \cellcolor{red!12}66.71 & 53.26 & \cellcolor{yellow!12}59.44 & \cellcolor{red!12}84.37 & \cellcolor{yellow!12}38.14 & \cellcolor{yellow!12}51.78 \\
    \textbf{CoverPrune-Lite}                        &                        & \cellcolor{yellow!12}59.43 & \cellcolor{yellow!12}68.87 & \cellcolor{yellow!12}54.03 & \cellcolor{yellow!12}66.31 & \cellcolor{red!12}57.64 & 57.18 & \cellcolor{yellow!12}82.20 & 37.11 & \cellcolor{red!12}52.10 \\
    \midrule

    VisionZip {\scriptsize (CVPR25)}       & \multirow{6}{*}{15\%} & 53.37 & 66.35 & 44.94 & 60.60 & 50.38 & 50.14 & 74.65 & \cellcolor{yellow!12}36.08 & 43.85 \\
    FastVID {\scriptsize (NeurIPS25)}      &                        & 54.03 & 66.51 & 47.13 & 61.67 & 52.29 & 49.44 & 72.12 & 35.05 & 48.06 \\
    DTC {\scriptsize (CVPR25)}             &                        & 55.07 & 68.41 & 49.00 & 63.99 & 45.28 & 49.01 & 76.99 & \cellcolor{red!12}38.66 & 49.19 \\
    EgoPrune {\scriptsize (arXiv25)}       &                        & 46.94 & 63.40 & 39.66 & 57.40 & \cellcolor{yellow!12}53.16 & 38.87 & 59.43 & 34.02 & 29.61 \\
    \textbf{CoverPrune}                             &                        & \cellcolor{red!12}58.27 & \cellcolor{red!12}69.42 & \cellcolor{red!12}55.10 & \cellcolor{red!12}66.46 & 52.81 & \cellcolor{red!12}58.03 & \cellcolor{yellow!12}77.13 & 35.57 & \cellcolor{red!12}51.62 \\
    \textbf{CoverPrune-Lite}                        &                        & \cellcolor{yellow!12}57.72 & \cellcolor{yellow!12}69.35 & \cellcolor{yellow!12}51.83 & \cellcolor{yellow!12}64.06 & \cellcolor{red!12}57.81 & \cellcolor{yellow!12}55.63 & \cellcolor{red!12}77.22 & 35.05 & \cellcolor{yellow!12}50.81 \\
    \midrule

    VisionZip {\scriptsize (CVPR25)}       & \multirow{6}{*}{10\%} & 50.36 & 64.74 & 41.80 & 58.58 & 49.55 & 42.39 & 66.83 & 34.02 & 44.98 \\
    FastVID {\scriptsize (NeurIPS25)}      &                        & 51.56 & 64.92 & 42.36 & 60.33 & 51.98 & 45.49 & 68.41 & 37.11 & 41.91 \\
    DTC {\scriptsize (CVPR25)}             &                        & 51.66 & 66.87 & 44.10 & 61.86 & 44.76 & 48.17 & 70.32 & 34.02 & 43.20 \\
    EgoPrune {\scriptsize (arXiv25)}       &                        & 44.71 & 63.06 & 35.10 & 53.99 & 52.88 & 37.46 & 54.73 & 34.54 & 25.89 \\
    \textbf{CoverPrune}                             &                        & \cellcolor{yellow!12}56.83 & \cellcolor{red!12}67.98 & \cellcolor{red!12}51.69 & \cellcolor{red!12}63.47 & \cellcolor{yellow!12}53.16 & \cellcolor{yellow!12}50.00 & \cellcolor{red!12}79.67 & \cellcolor{yellow!12}39.18 & \cellcolor{yellow!12}49.51 \\
    \textbf{CoverPrune-Lite}                        &                        & \cellcolor{red!12}56.94 & \cellcolor{yellow!12}67.13 & \cellcolor{yellow!12}47.91 & \cellcolor{yellow!12}63.19 & \cellcolor{red!12}55.76 & \cellcolor{red!12}53.10 & \cellcolor{yellow!12}78.43 & \cellcolor{red!12}39.69 & \cellcolor{red!12}50.32 \\
    \midrule

    VisionZip {\scriptsize (CVPR25)}       & \multirow{6}{*}{5\%}  & 46.10 & 62.87 & 36.38 & 54.59 & 51.11 & 36.76 & 54.68 & \cellcolor{red!12}37.11 & 35.28 \\
    FastVID {\scriptsize (NeurIPS25)}      &                        & 46.76 & 63.19 & 37.21 & 54.59 & \cellcolor{yellow!12}53.19 & 41.41 & 53.73 & 34.54 & \cellcolor{yellow!12}36.25 \\
    DTC {\scriptsize (CVPR25)}             &                        & 46.31 & 65.50 & 35.44 & 57.91 & 45.69 & 40.14 & 57.91 & 31.96 & 35.92 \\
    EgoPrune {\scriptsize (arXiv25)}       &                        & 40.85 & 62.02 & 29.93 & 52.40 & 52.19 & 30.85 & 47.79 & 32.99 & 18.61 \\
    \textbf{CoverPrune}                             &                        & \cellcolor{yellow!12}52.66 & \cellcolor{yellow!12}66.14 & \cellcolor{yellow!12}41.27 & \cellcolor{red!12}60.46 & 51.81 & \cellcolor{yellow!12}46.34 & \cellcolor{red!12}73.41 & \cellcolor{yellow!12}36.08 & \cellcolor{red!12}45.79 \\
    \textbf{CoverPrune-Lite}                        &                        & \cellcolor{red!12}52.88 & \cellcolor{red!12}66.30 & \cellcolor{red!12}44.45 & \cellcolor{yellow!12}60.19 & \cellcolor{red!12}54.69 & \cellcolor{red!12}47.75 & \cellcolor{yellow!12}68.82 & 35.05 & \cellcolor{red!12}45.79 \\
    \bottomrule
  \end{tabular}%
  }
\end{table}
\begin{table}[tb]
  \caption{Evaluation on VSI-Bench with VLM-3R\cite{vlm3r} as base model, where our methods outperform prior SOTA baselines on the average score and most individual tasks.}
  \label{tab:qa_style_table_vlm3r}
  \centering
  \small
  \setlength{\tabcolsep}{4pt}
  \renewcommand{\arraystretch}{1.15}
  \resizebox{\linewidth}{!}{%
  \begin{tabular}{@{}l c c cccccccc@{}}
    \toprule
    \multirow{2}{*}{Method} &
    \multirow{2}{*}{\makecell{Retention\\Ratio $R$}} &
    \multirow{2}{*}{\makecell{Avg.}} &
    \multicolumn{4}{c}{\cellcolor{orange!12}\textbf{Numerical Answer}} &
    \multicolumn{4}{c}{\cellcolor{yellow!12}\textbf{Multiple-Choice Answer}} \\
    & & &
    Obj.\ Count &
    Abs.\ Dist. &
    Obj.\ Size &
    Room Size &
    Rel.\ Dist. &
    Rel.\ Dir. &
    Route Plan &
    Appr.\ Order \\
    \midrule
    Vanilla & 100\% & 60.90 & 70.20 & 49.40 & 69.20 & 67.10 & 65.40 & 80.50 & 45.40 & 40.10 \\
    \midrule

    VisionZip {\scriptsize (CVPR25)}       & \multirow{6}{*}{10\%} & 50.76 & 63.47 & 40.14 & 62.82 & 50.52 & 53.38 & 61.41 & \cellcolor{yellow!12}44.85 & 29.45 \\
    FastVID {\scriptsize (NeurIPS25)}      &                        & 49.98 & 63.06 & 41.14 & 62.72 & 44.55 & 52.25 & 60.03 & \cellcolor{yellow!12}44.85 & 31.23 \\
    DTC {\scriptsize (CVPR25)}             &                        & 52.64 & 64.71 & \cellcolor{red!12}43.73 & 64.56 & \cellcolor{red!12}59.10 & 50.99 & 64.89 & 44.33 & 28.80 \\
    EgoPrune {\scriptsize (arXiv25)}       &                        & 47.49 & 62.80 & 39.17 & 61.49 & 48.96 & 44.65 & 57.97 & 44.33 & 20.55 \\
    CoverPrune                              &                        & \cellcolor{yellow!12}54.69 & \cellcolor{yellow!12}64.90 & \cellcolor{yellow!12}43.30 & \cellcolor{red!12}65.38 & 56.25 & \cellcolor{red!12}54.65 & \cellcolor{red!12}71.75 & \cellcolor{red!12}45.88 & \cellcolor{yellow!12}35.44 \\
    CoverPrune-Lite                         &                        & \cellcolor{red!12}54.74 & \cellcolor{red!12}65.54 & 43.20 & \cellcolor{yellow!12}64.91 & \cellcolor{yellow!12}56.94 & \cellcolor{yellow!12}54.51 & \cellcolor{yellow!12}70.24 & \cellcolor{yellow!12}44.85 & \cellcolor{red!12}37.70 \\
    \midrule

    VisionZip {\scriptsize (CVPR25)}       & \multirow{6}{*}{5\%}  & 46.30 & 62.16 & 37.13 & 61.41 & 40.31 & 50.42 & 50.10 & 42.78 & 26.05 \\
    FastVID {\scriptsize (NeurIPS25)}      &                        & 46.13 & 62.18 & 36.64 & 61.20 & 37.29 & 47.32 & 53.10 & \cellcolor{yellow!12}43.81 & 27.51 \\
    DTC {\scriptsize (CVPR25)}             &                        & 49.83 & \cellcolor{yellow!12}63.84 & \cellcolor{yellow!12}40.43 & 62.79 & \cellcolor{red!12}55.73 & 47.46 & 58.64 & \cellcolor{red!12}44.33 & 25.40 \\
    EgoPrune {\scriptsize (arXiv25)}       &                        & 44.22 & 61.86 & 36.24 & 60.19 & 42.40 & 42.25 & 49.26 & 43.30 & 18.28 \\
    \textbf{CoverPrune}                     &                        & \cellcolor{yellow!12}51.28 & 63.66 & 39.80 & \cellcolor{yellow!12}64.07 & 51.98 & \cellcolor{red!12}52.39 & \cellcolor{red!12}61.90 & 42.27 & \cellcolor{yellow!12}34.14 \\
    \textbf{CoverPrune-Lite}                &                        & \cellcolor{red!12}51.56 & \cellcolor{red!12}64.04 & \cellcolor{red!12}40.53 & \cellcolor{red!12}64.84 & \cellcolor{yellow!12}53.19 & \cellcolor{yellow!12}51.41 & \cellcolor{yellow!12}60.25 & 43.30 & \cellcolor{red!12}34.95 \\
    \bottomrule
  \end{tabular}%
  }
\end{table}
\subsubsection{Results on 3D Spatial Reasoning.}
Table~\ref{tab:qa_style_table_gs} and \ref{tab:qa_style_table_vlm3r} report results on VSI-Bench\cite{thinking}. Across both base models, CoverPrune and its Lite variant consistently set the highest average scores under matched token budgets, outperforming all baselines including attention-diversity fused strategies.

At 20\% token retention, CoverPrune preserves \textbf{92.4\%} of full-token performance. Our coverage-based paradigm’s advantage amplifies under aggressive pruning: both variants show far more graceful degradation than other SOTA methods, widening the performance gap as retention ratio drops, with substantially higher scores than the strongest competitor at 10\% and 5\% retention. It confirms that multi-domain coverage is critical for spatial reasoning under extreme compression. We further observe a complementary performance trend across the two variants: CoverPrune-Lite tends to perform better on global layout-focused Room Size tasks, while full CoverPrune shows advantages on fine-grained Relative Direction tasks. This aligns with their respective designs: CoverPrune-Lite’s geometry-aware ordering preserves coarse scale signals, while CoverPrune’s full coverage optimization retains fine-grained relational correspondences.
\begin{table}[tb]
  \centering
  \scriptsize
  \setlength{\tabcolsep}{3.5pt}
  \renewcommand{\arraystretch}{1.14}

  \begin{minipage}[t]{0.49\linewidth}
    \centering
    \captionof{table}{Ablation Study of CoverPrune with GS-Reasoner as the base model.}
    \label{tab:coverprune_ablation_r20}
    \resizebox{\linewidth}{!}{%
    \begin{tabular}{@{}l c c c c c@{}}
      \toprule
      Variant & \makecell[c]{$\Delta$Overall} & Overall & \makecell[c]{Rel.\\Dist.} & \makecell[c]{Rel.\\Dir.} & \makecell[c]{Appr.\\Order} \\
      \midrule
      CoverPrune                      & \textbf{0.00} & \textbf{59.76} & \textbf{59.44} & \textbf{84.37} & \textbf{51.78} \\
      w/o FST capacity                & -0.18         & 59.59          & 59.01          & 82.73          & 51.46 \\
      w/o feature cost       & -3.58         & 56.18          & 52.68          & 74.62          & 48.54 \\
      w/o geometry cost     & -0.74         & 59.02          & 59.15          & 81.20          & 50.65 \\
      w/o time cost         & -0.80         & 58.96          & 58.45          & 80.51          & 51.62 \\
      \bottomrule
    \end{tabular}%
    }
  \end{minipage}
  \hfill
  \begin{minipage}[t]{0.49\linewidth}
    \centering
    \captionof{table}{ Efficiency analysis of different pruning methods.}
    \label{tab:vsibench_eff_dtc_cp}
    \resizebox{\linewidth}{!}{%
    \begin{tabular}{@{}l c r r r r@{}}
      \toprule
      Methods & Tokens & \makecell[c]{Dec.\\Time\\(ms/token)} & \makecell[c]{Prun.\\Time\\(s)} & \makecell[c]{Memory\\(GB)} & \makecell[c]{Rel.\\Acc} \\
      \midrule
      \rowcolor{gray!15}
      Vanilla         & 6272 & 45.5 & 0.0    & 33.5 & 100.00 \\
      DTC             & 628  & \textbf{40.7} & 3.47 & \textbf{23.8} & 79.85 \\
      CoverPrune      & 628  & \textbf{40.7} & 2.53 & \textbf{23.8} & 87.84 \\
      CoverPrune-Lite & 628  & \textbf{40.7} & \textbf{0.41} & \textbf{23.8} & \textbf{88.01} \\
      \bottomrule
    \end{tabular}%
    }
  \end{minipage}
\end{table}
\subsection{Ablation and Analysis}
\subsubsection{Component Ablation.}
Table~\ref{tab:coverprune_ablation_r20} presents component ablation results on VSI-Bench ($R=20\%$), covering the FST capacity weighting and each term in the FST cost. Removing the feature cost causes the largest degradation, confirming semantic affinity is critical for fixed-budget coverage. Removing geometry or temporal cost incurs smaller but non-negligible drop, verifying 3D and temporal cues boost relation-centric reasoning. Disabling FST capacity weighting leads to a mild consistent decline, showing sufficient transport capacity benefits performance even with fixed retained tokens.

\subsubsection{Efficiency.}
Table~\ref{tab:vsibench_eff_dtc_cp} compares the efficiency of our method against the SOTA 3D VLM method DTC on VSI-Bench. Dec.\ Time, Prun.\ Time, and Rel.\ Acc denote decoding latency, pruning time overhead, and relative accuracy normalized to Vanilla, respectively. CoverPrune improves relative accuracy while reducing pruning overhead compared with DTC, and CoverPrune-Lite further cuts pruning time by a large margin while achieving the best relative accuracy and the lowest time cost. Since all pruned methods share the same memory reservation and decode latency at this budget, the speedup mainly comes from the lightweight coverage approximation in CoverPrune-Lite, which makes coverage-based selection more practical for real-time use.

\section{Conclusions and Future Directions}
In this work, we redefine 3D VLM visual token pruning as an OT-based coverage maximization problem, departing from the prevailing diversity-driven and attention-based paradigm. We propose CoverPrune, a training-free pruning framework built on our multi-domain FST transport cost, informativeness-aware FST target capacity, and an efficient SGS solver. We further introduce a lightweight variant, CoverPrune-Lite, which drastically reduces the pruning time overhead with minimal performance loss. Extensive experiments show our method consistently outperforms current SOTA baselines across benchmarks and base models. Looking ahead, we aim to comprehensively explore the full potential of this principled coverage-based pruning paradigm, extending its applicability to general VLMs, to deliver theoretically grounded, robust inference acceleration for universal large model deployment.

\subsubsection{Acknowledgements.}
This work was partly supported by the Special Foundations for the Development of Strategic Emerging Industries of Shenzhen
(No.~KJ{\allowbreak}ZD20231023094700001)
and the Shenzhen-Tsinghua Special Project for Fundamental \& Frontier Research in Artificial Intelligence
(No.~AI2026018).

\bibliographystyle{splncs04}
\bibliography{main}
\end{document}